\pdfoutput=1
\documentclass[11pt]{article}

\usepackage{ACL2023}
\usepackage{booktabs}
\usepackage{graphicx}
\usepackage{subfig}

\usepackage{times}
\usepackage{latexsym}
\usepackage{multirow}
\usepackage{amsmath}
\usepackage{amsfonts}

\usepackage[T1]{fontenc}
\usepackage[utf8]{inputenc}

\usepackage{microtype}

\usepackage{inconsolata}

\title{HiLight: A Hierarchy-aware Light Global Model with Hierarchical Local ConTrastive Learning}
\author{Zhijian Chen, Zhonghua Li, Jianxin Yang, Qi Ye \\
Huawei Technologies Ltd. Co. \\
\{chenzhijian13,lizhonghua3,yangjianxin4,ye.qi\}@huawei.com}

\begin{document}
\maketitle

\begin{abstract}
Hierarchical text classification (HTC) is a special sub-task of multi-label classification (MLC) whose taxonomy is constructed as a tree and each sample is assigned with at least one path in the tree. Latest HTC models contain three modules: a text encoder, a structure encoder and a multi-label classification head. Specially, the structure encoder is designed to encode the hierarchy of taxonomy. However, the structure encoder has scale problem. As the taxonomy size increases, the learnable parameters of recent HTC works grow rapidly. Recursive regularization is another widely-used method to introduce hierarchical information but it has collapse problem and generally relaxed by assigning with a small weight (ie. 1e-6). In this paper, we propose a \textbf{Hi}erarchy-aware \textbf{Li}ght \textbf{G}lobal model with \textbf{H}ierarchical local con\textbf{T}rastive learning (HiLight), a lightweight and efficient global model only consisting of a text encoder and a multi-label classification head. We propose a new learning task to introduce the hierarchical information, called Hierarchical Local Contrastive Learning (HiLCL). Extensive experiments are conducted on two benchmark datasets to demonstrate the effectiveness of our model.

\end{abstract}

\begin{figure}
\centering
	\includegraphics[width=\linewidth]{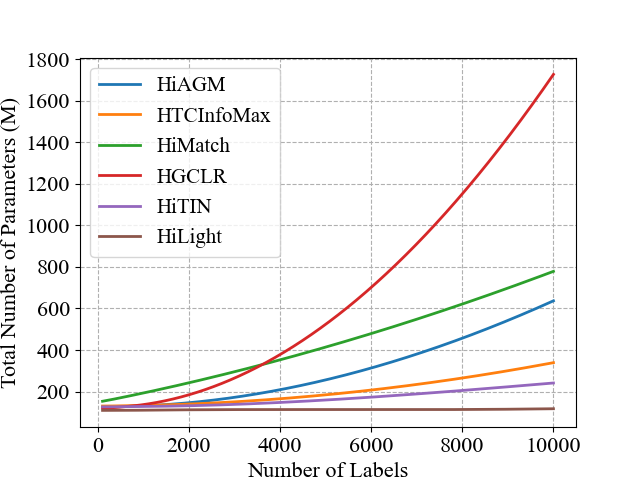}
	\caption{Learnable parameter size (in MB) of recent HTC works at taxonomy sizes. All models adopt BERT as text encoder.}
	\label{fig.growthRate}
\end{figure}

% 层次多标签分类任务介绍
% 1.突出与多标签分类的差异
% 2.引出利用层次信息必要性
\section{Introduction}
Hierarchical text classification (HTC) is a special sub-task of multi-label text classification whose taxonomy is constructed as a tree, commonly appears on structured databases like news \cite{lewis2004rcv1,Sandhaus2008nyt} and academic paper \cite{kowsari2017hdltex}. In HTC, each text is assigned with at least one path in the taxonomy tree. In order to make a more precise prediction, the hierarchical information should be taken into consideration during training or prediction, which is ignored in vanilla multi-label text classification. In a word, HTC is a kind of multi-label text classification task that make prediction based on textual and hierarchical information.

% HTC全局模型工作介绍
% 1.介绍现有工作如何利用层次信息
% 2.指出现有工作的不足 (参数量指数增长+推理慢)
Latest HTC models consist of 3 modules: a text encoder, a structure encoder and a multi-label classification head. The text encoder is used to encode textual feature and the structure encoder (typically GCN \citep{Thomas2017Semisupervised}) is designed to encode hierarchy. Recent researches focus on improving the network of structure encoder or the interaction between text encoder and structure encoder. \citet{jie2020hierarchy} is the first method to introduce structure encoder and proposes two interaction architectures between text encoder and structure encoder. \citet{deng-etal-2021-htcinfomax} proposes an information maximization task and \citet{chen-etal-2021-hierarchy} proposes two semantic matching tasks to improve text-label alignment and the label embedding learnt by structure encoder. \citet{zhu-etal-2023-hitin} proposes a tree isomorphism network to improve parameter efficiency of structure encoder. However, introducing a structure encoder will increase learnable parameters and have scaling problem. As the taxonomy size increases, the learnable parameters of recent HTC works grow rapidly as shown in Figure \ref{fig.growthRate}. Meanwhile, we doubt that is it necessary to introduce a clumsy structure encoder? On one hand, the taxonomy is tree-like and the connection is sparse \citep{jie2020hierarchy}. On the other hand, when the model is trained, the hierarchical information is static for different texts and the structure encoder is like a memory unit. \citet{wang-etal-2022-incorporating} proposes hierarchy-guided contrastive learning to remove structure encoder during inference by embedding the hierarchy into text encoder. However, \citet{wang-etal-2022-incorporating} still need a structure encoder to help generate contrastive samples during training. We wonder if there is a better way to introduce the hierarchical information without structure encoders?

% HTC不依赖结构工作介绍
% 1.介绍RR利用结构信息的方法
% 2.指出RR方法的不足 (坍塌问题)
Recursive regularization is another widely-used method to introduce the hierarchical information. \citet{gopal2013recursive} assumes that if two classifiers are close to each other in the parameter space, their behavior will be more consistent. Based on that assumption, \citet{gopal2013recursive} proposes the recursive regularization, which minimizes the distances between a classifier and its children classifiers in the parameter space. Recursive regularization tackles HTC just with a text encoder and a multi-label classification head, which can achieve rather good parameter efficiency. However, such regularization has transitivity and it may cause all classifiers collapse at some points in the parameter space. In order to maintain the discrimination, it is generally relaxed by assigning with a small weight, ie. 1e-6 \citep{jie2020hierarchy,chen-etal-2021-hierarchy,zhu-etal-2023-hitin}. We argue that such relaxation can't fix the above defect completely and a better learning task is still required.

% 提出本文工作
To this end, we propose Hierarchy-aware Light Global model with Hierarchical Local ConTrastive Learning (HiLight). HiLight only consists of a text encoder and a multi-label classification head, which can achieve good parameter efficiency. Besides, we propose a new learning task to introduce the hierarchical information, called Hierarchical Local Contrastive Learning (HiLCL). HiLCL consists of Local Contrastive Learning task (LCL) and Hierarchical Learning strategy (HiLearn), which are designed to enhance the discriminative ability of classifiers at same path in similar direction and consequently share similar behavior.

The main contribution of our work can be summarized as follows:

\begin{itemize}
    \item We propose HiLight model, a lightweight and efficient global model with a new effective learning task, which has no scaling problem.
    \item We propose HiLCL task, which enhances the discriminative ability of classifiers at same path in similar direction by proposed Local Contrastive Learning task (LCL) and Hierarchical Learning strategy (HiLearn).
    \item Extensive experiments are conducted on two benchmark datasets to demonstrate the effectiveness of our model.
\end{itemize}

\begin{figure*}
\centering
	\includegraphics[width=1.0\linewidth]{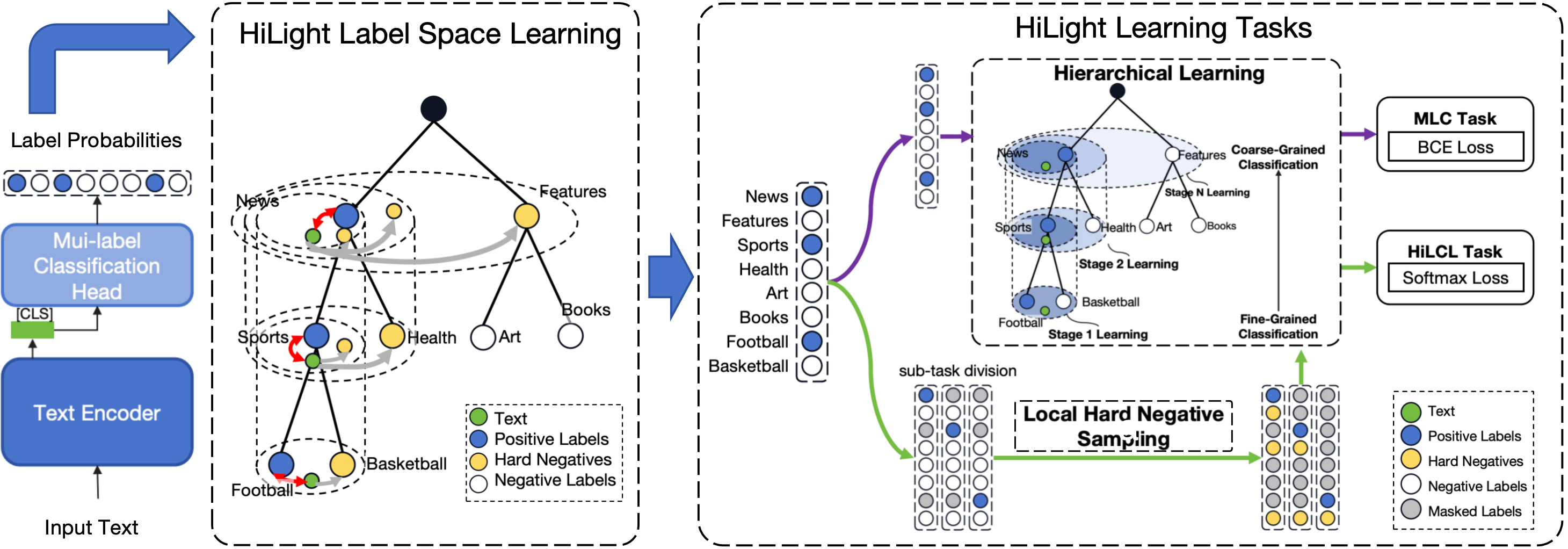}
	\caption{Illustration of HiLight. Given an input text, HiLight infers label probabilities by a text encoder and a multi-label classification head. With the inferred label probabilities and positive labels, HiLight conducts label space learning with MLC and HiLCL task. HiLCL is our proposed method and it divides the multi-label classification learning into multiple single-label classification learning. Then, HiLCL improves contrastive learning on each single-label classification learning with Local Hard Negative Sampling, which introduces negative labels from sibling and descendant label set of the positive label. Those negative labels outside the sibling and descendant label set are masked out during learning. HiLCL schedules learning with Hierarchical Learning strategy, which adopts a fine-to-coarse learning strategy to improve the discrimination of finest-grained labels.}
	\label{fig.hilight}
\end{figure*}

\section{Related Work}

% 相关工作介绍
% 1. 介绍层次分类任务及相关工作介绍
% 2. 介绍难负样本采样

\subsection{Hierarchical Text Classification}
Existing works in HTC could be categorized into local and global approaches. Local approaches build models for each node or each level in the hierarchy, which has parameter redundant problem. Meanwhile, global model treats HTC as a multi-label text classification task and only one model will be built, which has well parameter efficiency and becomes the mainstream. Traditional multi-label text classification treats each class independently and the activation of one class will not affect the other classes. However, classes in HTC are not independent. A class is predicted positive if at least one of its children is positive.

In order to improve behavioral consistency along a path in the hierarchy, early works try to constrain classifiers via regularization (\citet{gopal2013recursive}), reinforcement learning (\citet{yuning2019hierarchical}) and meta-learning (\citet{jiawei2019learning}). Later on, \citet{jie2020hierarchy} introduces structure encoder to encode the hierarchy of taxonomy and proposes two interaction architecture between text encoder and structure encoder. Following \citet{jie2020hierarchy}, recent works focus on improving the network of structure encoder or the interaction between text encoder and structure encoder. \citet{deng-etal-2021-htcinfomax} proposes an information maximization task and \citet{chen-etal-2021-hierarchy} proposes two semantic matching tasks to improve text-label alignment and the label embedding learnt by structure encoder. \citet{zhu-etal-2023-hitin} proposes a tree isomorphism network to improve parameter efficiency of structure encoder.

\subsection{Hard Negative Sampling}\label{hard_neg_sample}
Negative sampling is a commonly used machine learning technique \citep{mikolov2013distributed,mao2021boosting,Galanopoulos2021Hard}, which reduces computational costs and accelerates convergence speed by sampling a subset of negative samples. Different negative sampling methods adopt its own strategy. Hard negative sampling is the one that adopts the strategy of mining the negative examples closest to the anchor or positive examples. These negatives are the most confusing for the model and drive the model to improve its discriminative ability.

In HTC, \citet{chen-etal-2021-hierarchy} also introduces negative sampling to select negative labels. For each positive label, \citet{chen-etal-2021-hierarchy} select its parents, one of siblings and a random label from the hierarchy as negative label set. Our proposed method is different from \citet{chen-etal-2021-hierarchy} in three ways. Firstly, we do not introduce a structure encoder to generate label representations; Secondly, we adopt softmax loss instead of margin loss as the loss function, which can see more negatives at one training step; Lastly, for each positive label, we select all of its negative siblings and negative descendant labels as the negative label set. Since the hierarchy of HTC is tree-like, classifiers on the same path will share similar hard negatives, which will drive them to have similar discriminative ability.

\section{Problem Definition}\label{definition}

Given an input text $x = \{x_1, x_2, \ldots, x_n\}$, hierarchical text classification (HTC) aims to predict a subset $y=\{y_1, y_2, \ldots, y_m\}$ from the label set of taxonomy $Y = \{y_1, y_2, \ldots, y_C\}$, where $n$ donates the text length, $m$ donates the size of ground truth label subset and $C$ donates the taxonomy size.  The taxonomy is predefined and organized as an acyclic graph $G=(Y, E)$, where $E$ is the set of edge connecting a node and its parents. Since a non-root node has only one parent, the taxonomy graph can be converted to a tree. When the max depth of taxonomy tree reduces to one, the HTC will be degraded to the vanilla multi-label classification (MLC). Elements of $y$ come from at least one path in $G$. Behavioral consistency is defined as follows: a class is predicted as positive if at least one of its children is positive; and negative if all its children are negative. Formally,

\begin{equation}
\label{equ:hiaware}
\left\{
\begin{array}{ll}
    P(y_i) \geq \theta, & \exists e_{ij} \in E, P(y_j) \geq \theta \\
    P(y_i) <    \theta, & \forall e_{ij} \in E, P(y_j) <    \theta \\
\end{array}
\right.
\end{equation}
where $\theta$ is the activation threshold.

\section{Hierarchy-aware Light Global Model}

As depicted in Figure \ref{fig.hilight}, we propose the \textbf{Hi}erarchy-aware \textbf{Li}ght \textbf{G}lobal model with \textbf{H}ierarchical local con\textbf{T}rastive learning (HiLight). HiLight consists of a text encoder and a multi-label classification head. Meanwhile, we propose \textbf{Hi}erarchical \textbf{L}ocal \textbf{C}ontrastive \textbf{L}earning (HiLCL) task to introduce hierarchical information. Details will be described as follows.

\begin{figure*}
\centering
	\includegraphics[width=1.0\linewidth]{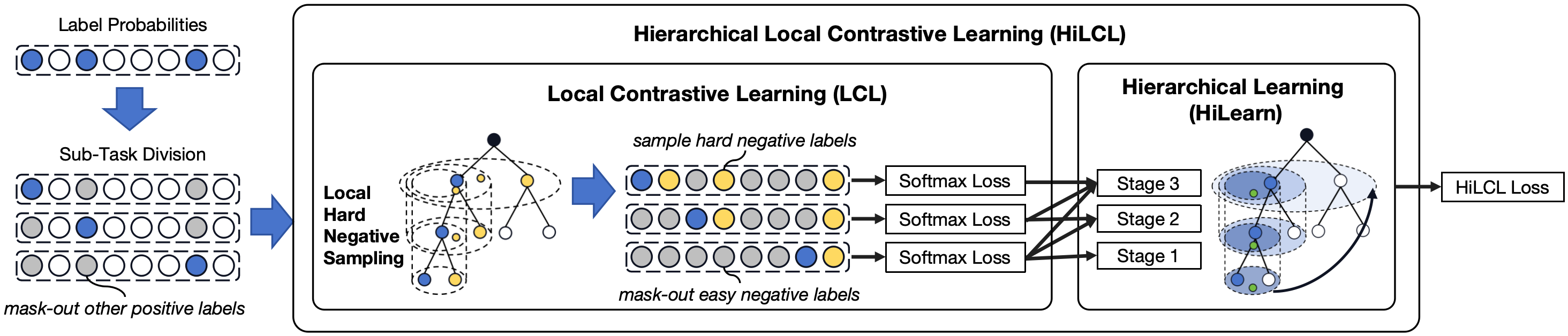}
	\caption{An Example of HiLCL with target size of 3. Firstly, HiLCL divides the multi-label classification learning into 3 single-label classification learning. For each positive label, HiLCL conducts LCL task, which masks out outputs of other positive labels as well as easy negative labels and then contrasts the output of current positive label with outputs of hard negative labels. Meanwhile, HiLCL schedules the LCL learning with HiLearn, which learns finest-grained positive labels at early epochs and adds coarse-grained positive labels gradually.}
	\label{fig.hilcl}
\end{figure*}

\subsection{Text Encoder}

TextRCNN\citep{lai2015recurrent} and BERT\citep{devlin-etal-2019-bert} are two popular text encoders adopted by recent HTC works \citep{jie2020hierarchy,deng-etal-2021-htcinfomax,chen-etal-2021-hierarchy,wang-etal-2022-incorporating,zhu-etal-2023-hitin}. BERT is a pre-trained model that absorbs rich common knowledge and hence masters strong semantic understanding ability. Recent state-of-the-art results\citep{deng-etal-2021-htcinfomax, chen-etal-2021-hierarchy, wang-etal-2022-incorporating, zhu-etal-2023-hitin} are achieved by adopting BERT as text encoder. In order to build a lightweight and efficient model, we need a strong text encoder and thus we adopt BERT as text encoder. Given an input text:
\begin{equation}
    \label{equ:inputs}
    x = \{[CLS], x_1, x_2, \ldots, x_{n}, [SEP]\}
\end{equation}
where [CLS] and [SEP] are two special tokens indicating the beginning and the end of the sequence. The length of the sequence will be taken as $n$ for convenience. The text encoder maps each token into a hidden state:
\begin{equation}
    \label{equ:enc}
    H = BERT(x)
\end{equation}
where $H \in \mathbb{R}^{n\times d_h}$ and $d_h$ is the hidden state size. Among all hidden states, the one of [CLS] has the best awareness of global context. Therefore, we take it as the global hidden state of the input text $h=H_{[CLS]}$.

\subsection{Multi-label Classification Head}

The multi-label classification head performs a mapping from hidden state space to label space as shown in Figure \ref{fig.hilight}. The distance between the input text and a class in label space will be regarded as label probability. We adopt a linear layer with a sigmoid activation function as the classification head. We also add dropout to the hidden state to improve generalization like previous works \citep{jie-etal-2022-learning,zhu-etal-2023-hitin}:
\begin{equation}
    \label{equ:cls_head}
    P = \sigma (W \cdot Dropout(h)\cdot + b)
\end{equation}
where $W \in \mathbb{R}^{C\times d_h}$ and $b \in \mathbb{R}^{C}$. The predicted label set $y$ is a set of labels whose probabilities are greater than the predefined threshold $\theta$:
\begin{equation}
    \label{equ:cls}
    y = \{y_i | P_i \geq \theta, y_i \in Y\}
\end{equation}
where $P_i$ is the i-th output of P.

\subsection{Hierarchical Local Contrastive Learning}

% The key of label probabilities inference is maintaining behavioral consistency as defined in section \ref{definition}. Previous works encode the hierarchy with structure encoder while we propose to use the hierarchy to mine local hard negative labels. Here we propose a new learning task, called Hierarchical Local Contrastive Learning task (HiLCL) shown in Figure \ref{fig.hilcl}. HiLCL conducts Local Contrastive Learning (LCL) with proposed Local Hard Negative Sampling, which is designed to guide classifiers at same path to have similar discriminative ability.

% Meanwhile, the behavior of classifiers at same path is also coarse-to-fine label space division. The label spaces at lower level are sub-spaces of those at higher level. Since LCL adopts hard negative labels from lower levels, the label spaces of those finer-grained negative labels will be affected due to LCL punishment. In order to maintain the discrimination of finer-grained labels, we propose Hierarchical Learning (HiLearn) to schedule the learning of LCL, which adopts a fine-to-coarse learning strategy.

In HTC, the key of label probabilities inference is maintaining behavioral consistency as defined in section \ref{definition}. Previous works encode the hierarchy with structure encoder while we propose to use the hierarchy to mine local hard negative labels. Here we propose a new learning task, called Hierarchical Local Contrastive Learning task (HiLCL) as shown in Figure \ref{fig.hilcl}. HiLCL consists of Local Contrastive Learning task (LCL) and Hierarchical Learning strategy (HiLearn).

\subsubsection{Local Contrastive Learning}

Contrastive learning is a powerful representative space learning task in CV \citep{he2020momentum,chen2020simple} and NLP \citep{gao2021simcse,kim2021self}. The target of contrastive learning is to encourage the distances of positive pairs smaller than the distances to a bunch of negative samples. Similarly, the goal of HTC is to make the input text close to the positive labels and far away from the negative labels in the label space. The difference is that in HTC, the positive labels are hierarchically related and their behavior should be consistent with their relationship. Thus, we propose Local Contrastive Learing task (LCL), which improves the consistency of contrastive learning by Local Hard Negative Sampling.

\textbf{Local Hard Negative Sampling}. As mentioned in section \ref{hard_neg_sample}, hard negative sampling adopts a strategy of sampling the nearest negative samples, which are challenging for the model. In HTC, given an input text and one of its positive labels, the closet labels are the sibling and descendant of the positive label as shown in Figure \ref{fig.hilcl}. Thus, we propose to sample the negative sibling and negative descendant labels as the hard negative label set, called Local Hard Negative Sampling. Intuitively, these negative labels are challenging because they are different aspects of same category. They share common characteristics with the positive label but there are also subtle differences; On the contrary, those negative labels outside the sibling and descendant set, share less characteristics with the positive label and easier to distinguish, which are treated as easy negative labels and masked out during learning as shown in Figure \ref{fig.hilcl}. Formally,

\begin{tiny}
\begin{equation}
\label{equ:hardneg}
\begin{aligned}
Hard(y_i) = \{v_j | v_j \in Sibiling(y_i) \cup SubTree(y_i), v_j \notin y\}
\end{aligned}
\end{equation}
\end{tiny}
\begin{small}
\begin{equation}
\label{equ:sibling}
\begin{aligned}
Sibiling(y_i) = \{v_j | \exists v_k \in Y, e_{ki} \in E, e_{kj} \in E\}
\end{aligned}
\end{equation}
\end{small}
\begin{small}
\begin{equation}
\label{equ:subtree}
\begin{aligned}
SubTree(y_i) = \{v_j | \forall v_k \in (y_i, \ldots, v_j), e_{kk+1} \in E \}
\end{aligned}
\end{equation}
\end{small}
where $Hard(y_i)$, $Sibling(y_i)$ and $SubTree(y_i)$ are the hard negative label set, the sibling label set and the descendant label set of $y_i$ correspondingly.

Combining contrastive learning and Local Hard Negative Sampling, we propose Local Contrastive Learning task (LCL). Formally,

\begin{tiny}
\begin{equation}
\label{equ:lcl}
\begin{aligned}
L_{LCL}(y_i) & = - \log \frac{exp(P_{i})}{exp(P_{i}) + \sum_{v_j \in Hard(y_i)} exp(P_{j})} \\
               = - \log & \frac{exp(\sigma (W_i h + b_i))}{exp(\sigma (W_i h + b_i)) + \sum_{v_j \in Hard(y_i)} exp(\sigma (W_j h + b_j))} \\
            \sim - \log & \frac{exp(W_i h + b_i)}{exp(W_i h + b_i) + \sum_{v_j \in Hard(y_i)} exp(W_j h + b_j)} \\
\end{aligned}
\end{equation}
\end{tiny}
where $W_i$ and $b_i$ are the weight of classifier $i$.

Since the positive labels are located within same sub-space, their hard negative labels are largely overlapped. Conducting LCL on the corresponding classifiers, their discriminative ability will be enhanced in similar direction and consequently share similar behavior.

\subsubsection{Hierarchical Learning}

Hierarchical Learning is the scheduling strategy for LCL. Intuitively, HTC is a coarse-to-fine classification process and correspondingly the behavior of classifiers at same path is a coarse-to-fine label space division process. The scopes of label spaces along any path degrade gradually and eventually the label spaces at leaf level are the finest-grained division of the whole space. LCL adopts negative labels from lower levels and the space division learning of finer-grained labels will be affected due to suppression of softmax loss in Eq.\ref{equ:lcl}.

In order to improve the discrimination of finest-grained labels, we propose a fine-to-coarse learning strategy, called Hierarchical Learning (HiLearn). For each training sample ($x$, $y$), HiLearn samples a subset of $y$ as the target set at each epoch $ep$ and enlarges the target set every $k$ epoch by adding labels from higher levels. Formally,

\begin{small}
\begin{equation}
\label{equ:hilearn}
\begin{aligned}
HiLearn(y, ep) & = \{y_i|D_{rev}(y_i) \leq \lfloor ep / k \rfloor, y_i \in y\} \\
\end{aligned}
\end{equation}
\end{small}
where $D_{rev}(\cdot)$ is the reverse depth function starting from leaf and the depth of a leaf node is 0 and $k \in N$ is the scheduling parameter of HiLearn.

\subsubsection{Hierarchical Local Contrastive Learning}

Combining LCL and HiLearn, we propose Hierarchical Local Contrastive Learning task (HiLCL). HiLCL divides the multi-label classification learning into multiple single-label classification learning as shown in Figure \ref{fig.hilcl}. Then, HiLCL conducts LCL on each classifier (Eq.\ref{equ:lcl}) and schedules the learning with HiLearn (Eq.\ref{equ:hilearn}), which adopts a fine-to-coarse strategy. Formally,

\begin{equation}
\label{equ:hilcl}
\begin{aligned}
L_{HiLCL}(y, ep) & = \sum_{y_i \in HiLearn(y, ep)} L_{LCL}(y_i) \\
\end{aligned}
\end{equation}

\subsection{Multi-label Classification Learning}

Besides HiLCL, we also adopt MLC task to instruct model learning. We use the Binary Cross-Entropy loss (BCE) as loss function. Formally,

\begin{equation}
\label{equ:bce}
\begin{aligned}
L_{BCE}(y) &= - \frac{1}{|y|} \sum_{y_i \in y} log(P_i) \\
           &= - \frac{1}{|y|} \sum_{y_i \in y} log \frac{exp(W_i h + b_i)}{1 + exp(W_i g + b_i)}
\end{aligned}
\end{equation}
Thus, the final loss function can be formulated as:
\begin{equation}
\label{equ:loss}
L = L_{BCE} + \lambda \cdot L_{HiLCL}
\end{equation}
where $\lambda$ is the balancing factor.

\begin{table}\small
\setlength{\tabcolsep}{1.6pt}
\centering  
\begin{tabular}{lcccccc}
\toprule
Dataset & |Y| & $Avg(|y_i|)$ & Depth & Train & Dev & Test \\
\hline
WOS     & 141 & 2.0 & 2 & 30,070 & 7,518 & 9,397 \\
RCV1-v2 & 103 & 3.24 & 4 & 20,833 & 2,316 & 781,265 \\
\bottomrule
\end{tabular}
\caption{Summary statistics of datasets}
\label{tab:dataset}
\end{table}

\begin{table*}[t]\small
\centering  
\begin{tabular}{lcccccc} 
\toprule
\multirow{2}{*}{Methods} & \multicolumn{2}{c}{WOS} &  & \multicolumn{2}{c}{RCV1-v2} \\ 
\cmidrule{2-3}\cmidrule{5-6}
 & Micro-F1 & Macro-F1 &  & Micro-F1 & Macro-F1 \\
\hline
HiAGM$^\dagger$ \citep{jie2020hierarchy}   & 86.04 & 80.19  &  & 85.58 & 67.93 \\
HTCInfoMax$^\dagger$ \citep{deng-etal-2021-htcinfomax}  & 86.30 & 79.97 &  & 85.53 & 67.09 \\
HiMatch$^\dagger$ \citep{chen-etal-2021-hierarchy}  & 86.70 & 81.06 &  & 86.33 & 68.66 \\
HGCLR$^\dagger$ \citep{wang-etal-2022-incorporating}  & 87.11 & 81.20 &  & 86.49 & 68.31 \\
HiTIN$^\dagger$ \citep{zhu-etal-2023-hitin}  & 87.19 & 81.57 &  & 86.71 & \textbf{69.95} \\
\hline
HiLight (Ours) & \textbf{87.63} & \textbf{82.36} &  & \textbf{86.89} & 69.58 \\ 
\bottomrule
\end{tabular}
\caption{Experimental results of out proposed method on several datasets. All baselines adopt BERT as the text encoder. $\dagger$ donates the results reported by \citeposs{zhu-etal-2023-hitin}.}
\label{tab:classification}
\end{table*}

\section{Experiment}

\subsection{Experiment Setup}

\textbf{Datasets and Evaluation Metrics} Experiments are conducted on Web-of-Science (WOS) \citep{kowsari2017hdltex} and RCV1-v2 \citep{lewis2004rcv1} for comparison and analysis. WOS makes up of abstracts of published papers from Web-of-Science while RCV1-v2 consists of news passages. We follow the data processing and data partitioning of HiAGM\footnote{https://github.com/Alibaba-NLP/HiAGM} \cite{jie2020hierarchy}. The statistic details of these two datasets are shown in Table \ref{tab:dataset}. WOS is for the single-path HTC while RCV1-v2 includes multi-path taxonomic tags. WOS is wide and shallow while RCV1-v2 is deeper. Both datasets are representative scenarios in real-world applications. We adopt Micro-F1 and Macro-F1 as evaluation metrics like previous works.

\noindent \textbf{Implementation Details} We adopt \texttt{bert-base-uncased}\footnote{https://huggingface.co/bert-base-uncased} from Huggingface as text encoder. The hidden state of [CLS] is taken as the global hidden state of input text. Dropout is applied on top of the global hidden state and dropout rate is 0.1. We adopt Adam as optimizer. The learning rate of text encoder is 2e-5 while the classification head is 1e-3. For HiLCL, the balancing factor $\lambda$ is 1e-2 for WOS and 1e-3 for RCV1-v2. For HiLearn, the scheduling parameter $k$ is 3 for WOS and 2 for RCV1-v2. We train the model for 100 epochs and decay the learning rate with 0.8 if the Micro-F1 or Macro-F1 on dev does not increase for 5 epochs. In order to ensure reproducibility, we configure the random seed with 2023. Training is conducted on a NVIDIA P100 GPU and takes about 2 days for WOS and 1 day for RCV1-v2.

\noindent \textbf{Baselines} We choose some recent HTC global models as baselines, including HiAGM \citep{jie2020hierarchy}, HTCInfoMax \citep{deng-etal-2021-htcinfomax}, HiMatch \citep{chen-etal-2021-hierarchy}, HGCLR \citep{wang-etal-2022-incorporating} and HiTIN \citep{zhu-etal-2023-hitin}. HiAGM introduces the structure encoder and proposes two label-text interaction architectures. HTCInfoMax proposes the information maximization task to enhance HiAGM by regularizing the label representation with a prior distribution. HiMatch proposes two embedding matching tasks to improve label-text alignment in a joint embedding space. HGCLR proposes to incorporate hierarchy into text encoder with contrastive learning and infer labels without structure encoder. HiTIN proposes a tree isomorphism network to improve the parameter efficiency of structure encoder. All approaches adopt BERT as text encoder for a fair comparison.

\subsection{Experiment Results}
The experimental results are shown in Table \ref{tab:classification}. Considering the scaling problem of structure encoder, we rethink the key of HTC and instead of introducing a structure encoder to encode hierarchy directly like baselines, we propose HiLCL task that uses the hierarchy to mine local hard negative labels without introducing extra parameters. By analysing previous works, we believe that it is possible to incorporate the hierarchical information with a simple learning task and the experimental results support our view. In both WOS and RCV1, our method show competitive performance comparing with baselines. Especially in WOS, our method beats all structure-encoder-based methods, which shows the effectiveness of our method. In RCV1-v2, we only fail to beat HiTIN on Macro-F1. 

HiAGM is the first model to introduce a GCN structure encoder and establishes a baseline for global model. HTCInfoMax and HiMatch enhance HiAGM by introducing new learning tasks. The best result of introducing a GCN structure encoder is achieved by HiMatch. HiTIN proposes a new structure encoder, called tree isomorphism network. HiTIN beats HiMatch and introduces much less learnable parameters than HiMatch. HGCLR make a further improvement by removing structure encoder completely during inference and also beats HiMatch, which shows the possibility of introducing the hierarchical information without the structure encoder. However, HGCLR still needs a structure encoder during training and fails to beat HiTIN, which shows that there are rooms for improvement. Our model introduce the hierarchical information with the simplest model and a more effect learning task, which beats HGCLR and HiTIN (excpet for Macro-F1 on RCV1-v2) and proves that structure encoders is not the only way.

\subsection{Analysis}

% \begin{table}\small
% \setlength{\tabcolsep}{1pt}
% \centering
% \begin{tabular}{lcccccc}
% \toprule
% \multirow{2}{*}{Ablation Models} & \multicolumn{2}{c}{WOS} & & \multicolumn{2}{c}{RCV1-v2} \\ 
% \cmidrule{2-3}\cmidrule{5-6}
%  & Micro-F1 & Macro-F1 &  & Micro-F1 & Macro-F1 \\ 
% \hline
% BERT & 86.60 & 80.71 & & 86.41 & 68.84 \\
% BERT+HiTIN & 86.98 & 81.58 & & 86.88 & 69.21 \\
% BERT+HiLight & 86.40 & 80.51 & & 86.19 & 67.22 \\
% \hline
% BERT$^\ddagger$ & 87.30 & 81.93 & & 86.86 & 69.05 \\
% BERT$^\ddagger$+HiTIN & 86.69 & 81.16 & & \textbf{87.08} & \textbf{69.59} \\
% BERT$^\ddagger$+HiLight & \textbf{87.63} & \textbf{82.36} & & 86.79 & 69.39 \\
% \bottomrule
% \end{tabular}
% \caption{Ablation study on model settings.}
% \label{tab:ablate_bert}
% \end{table}

\begin{table}
\setlength{\tabcolsep}{1pt}
\centering
\begin{tabular}{lcccccc}
\toprule

% \multirow{2}{*}{Ablation Models} & \multicolumn{2}{c}{WOS} & & \multicolumn{2}{c}{RCV1-v2} \\ 
% \cmidrule{2-3}\cmidrule{5-6}
%  & Micro-F1 & Macro-F1 &  & Micro-F1 & Macro-F1 \\ 
% \hline
% HiLight & \textbf{87.63} & \textbf{82.36} & & 86.89 & \textbf{69.58} \\
% - \textit{r.p.} coarse-to-fine &  &  & &  &  \\
% - \textit{r.m.} fine-to-coarse & 87.36 & 81.94 & & \textbf{86.95} & 69.08 \\
% - \textit{r.p.} random negatives &  &  & &  &  \\
% - \textit{r.m.} sibling negatives &  &  & &  & \\
% - \textit{r.m.} subtree negatives &  &  & &  & \\
% - \textit{r.m.} HiLCL & 87.30 & 81.93 & & 86.86 & 69.05 \\

Ablation Models & Micro-F1 & Macro-F1 \\ 
\hline
HiLight & 86.89 & \textbf{69.58} \\
- \textit{r.p.} coarse-to-fine    & 86.32 & 66.23 \\
- \textit{r.m.} fine-to-coarse    & \textbf{86.95} & 69.08 \\
- \textit{r.p.} random negatives  & 86.80 & 68.69 \\
- \textit{r.m.} sibling negatives & 86.86 & 69.23 \\
- \textit{r.m.} subtree negatives & 86.89 & 69.15 \\
\bottomrule
\end{tabular}
\caption{Ablation study on the components of HiLight. \textit{r.p.} stands for \textit{replace} and \textit{r.m.} stands for \textit{remove}. Experiments are conducted on RCV1-v2.}
\label{ablate_hilight}
\end{table}

\subsubsection{Ablation Study}

In this section, we investigate to study the independent effect of each component in our proposed model on RCV1-v2 dataset. The results are reported in Table \ref{ablate_hilight}.

Firstly, we validate the influence of HiLearn strategy. After replacing HiLearn strategy with coarse-to-fine strategy, we can see a large drop on Micro-F1 and Macro-F1. We assume that learning coarse-grained classification initially will make the text encoder biased on some coarse-grained features. After that, further learning is based on previous learnt features, which may cause overfitting. Besides, we also experiment without HiLearn strategy and we can see a drop on Macro-F1 but a raise on Micro-F1, which indicates that the model may converge to some labels with more training data. Above results show that introducing HiLearn can improve the discrimination of labels with less training data.

Secondly, we validate the influence of Local Negative Sampling. By removing sibling negatives or subtree negatives or replacing them with random negatives, we can see a drop on Micro-F1 and Macro-F1, which shows that both sibling negatives and subtree negatives are necessary for LCL.

\begin{figure}
\centering
	\includegraphics[width=\linewidth]{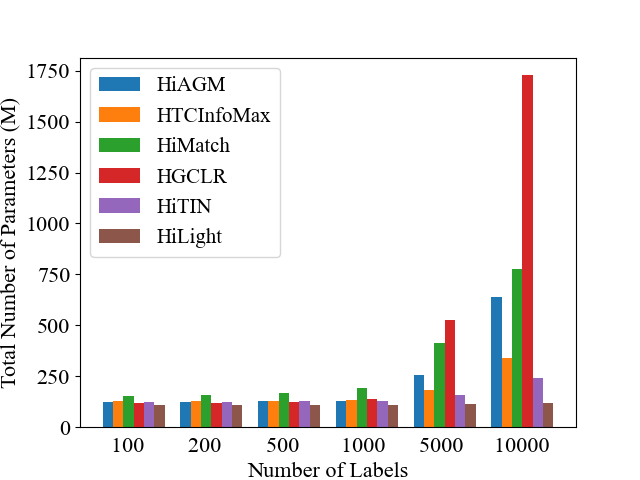}
	\caption{Learnable parameter size of recent HTC models at different taxonomy sizes. All models adopt BERT as text encoder.}
	\label{paramertersbar}
\end{figure}

\subsubsection{Parameter Efficiency Study}

In this section, we investigate the learnable parameter size of recent HTC works at different taxonomy sizes. As shown in Figure \ref{paramertersbar}, when the taxonomy size increases from 100 to 10000, the learnable parameter size of HiLight basically remains unchanged. However, recent HTC works (\citep{jie2020hierarchy,deng-etal-2021-htcinfomax,chen-etal-2021-hierarchy,wang-etal-2022-incorporating,zhu-etal-2023-hitin}) increase at different degrees. Among all, HGCLR \citep{wang-etal-2022-incorporating} has the largest growth rate, which increases from 121 (MB) to 1,728 (MB). Although HGCLR removes the structure encoder during inference like us, their training cost is much higher than us.

\begin{figure*}[htbp]
\centering
\subfloat[HiLight(Ours)]
{
    \label{fig:subfig1}
    \includegraphics[width=0.3\linewidth]{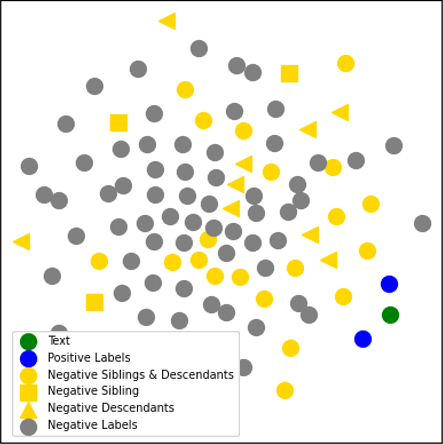}
}
\subfloat[Recursive Regularization]
{
    \label{fig:subfig2}
    \includegraphics[width=0.3\linewidth]{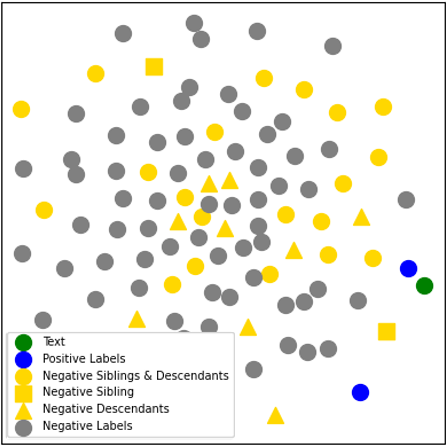}
}
\subfloat[BERT]
{
    \label{fig:subfig3}
    \includegraphics[width=0.3\linewidth]{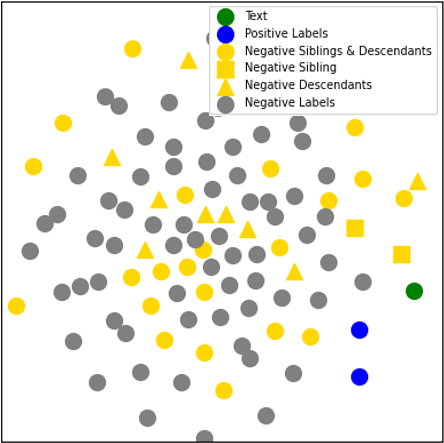}
}
\caption{T-SNE visualization of label space mapping on RCV1-v2. Green dots indicate the input text. Blue dots indicate positive labels. Yellow dots indicate negative sibling and descendant labels. Yellow triangles indicate negative descendant labels. Yellow squares indicate negative sibling labels. Grey dots indicate easy negative labels.}
\label{fig:subfig_1}
\end{figure*}

\begin{table}
\setlength{\tabcolsep}{1pt}
\centering
\begin{tabular}{lcccccc}
\toprule

% \multirow{2}{*}{Methods} & \multirow{2}{*}{Weight} & \multicolumn{2}{c}{WOS} & & \multicolumn{2}{c}{RCV1-v2} \\ 
% \cmidrule{3-4}\cmidrule{6-7}
% & & Micro-F1 & Macro-F1 &  & Micro-F1 & Macro-F1 \\ 
% \hline
% Rec$^\ddagger$ & 1e-2 &  &  & &  &  \\
% Rec$^\ddagger$ & 1e-3 &  &  & & 86.60 & 60.42 \\
% Rec$^\ddagger$ & 1e-6 & 87.19 & 81.92 & & \textbf{87.03} & 68.76 \\
% \hline
% HiLight & 1e-2 & \textbf{87.63} & \textbf{82.36} & & 86.85 & 69.34 \\
% HiLight & 1e-3 &                &                & & 86.89 & \textbf{69.58} \\
% HiLight & 1e-6 &                &                & &       &  \\

Models & Weight & & Micro-F1 & Macro-F1 \\
\hline
\multirow{3}{*}{Rec Reg} & 1e-2 & & 84.21 & 38.72 \\
 & 1e-3 & & 86.60 & 60.42 \\
 & 1e-6 & & 87.03 & 68.76 \\
\hline
\multirow{3}{*}{HiLight} & 1e-2 & & 86.85 & 69.34 \\
 & 1e-3 & & 86.89 & 69.58 \\
 & 1e-6 & & 86.91 & 69.14 \\

\bottomrule
\end{tabular}
\caption{Effect of encoding hierarchy without structure encoders. Rec Reg is Recursive Regularization  \citep{gopal2013recursive} for short. Experiments are conducted on RCV1-v2.}
\label{ablate_reg}
\end{table}

\subsubsection{Collapse Problem Study}

\begin{figure}[htbp]
\centering
\subfloat[Rec Reg]
{
    \label{fig:subfig11}
    \includegraphics[width=0.5\linewidth]{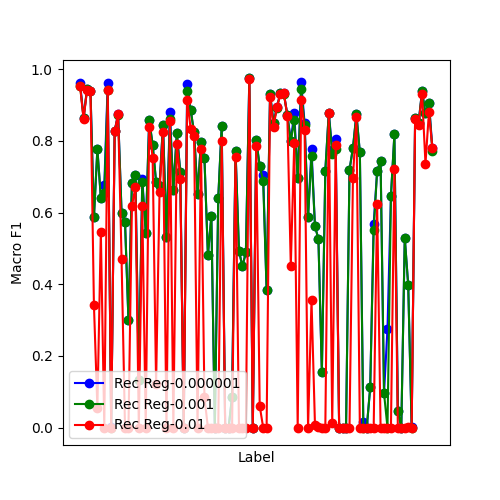}
}
\subfloat[HiLight (Ours)]
{
    \label{fig:subfig22}\includegraphics[width=0.5\linewidth]{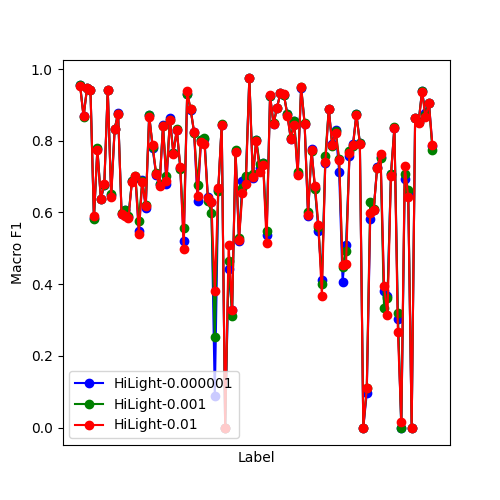}
}
\caption{Label-wise MacroF1 on RCV1-v2. The abscissa indicates the labels and ordinate indicates the Macro-F1. Red lines indicate weight=1e-2. Green lines indicate weight=1e-3. Blue lines indicate weight=1e-6.}
\label{fig:labelwiseMacroF1}
\end{figure}

We claim that the collapse problem is the main problem of recursive regularization. In this section, we verify our assumption by varying the weight of learning tasks. Experiments are conducted on RCV1-v2 dataset and the results are shown in Table \ref{ablate_reg}. We can see that recursive regularization is sensitive to the weight. As the weight increases, Micro-F1 drops slowly while Macro-F1 falls down dramatically. We plot label-wise Macro-F1 on Figure \ref{fig:labelwiseMacroF1} and we can see that Macro-F1 of some labels drop to near zero while other remain nearly unchanged, which indicates that the model may convergence to some labels with more training data. Meanwhile, the results of our model are much steadier. As the weight changes, both Micro-F1 and Macro-F1 fluctuate within a small range. Above results verify our assumption and demonstrates the effectiveness of our model.

% \begin{table}\small
% \setlength{\tabcolsep}{1pt}
% \centering
% \begin{tabular}{lcccccc}
% \toprule
% \multirow{2}{*}{Models} & \multicolumn{2}{c}{WOS} & & \multicolumn{2}{c}{RCV1-v2} \\ 
% \cmidrule{2-3}\cmidrule{5-6}
%  & Micro-F1 & Macro-F1 &  & Micro-F1 & Macro-F1 \\ 
% \hline
% Rec$^\ddagger$ (1e-2) &  &  & &  &  \\
% Rec$^\ddagger$ (1e-3) &  &  & & 86.60 & 60.42 \\
% Rec$^\ddagger$ (1e-6) & 87.19 & 81.92 & & \textbf{87.03} & 68.76 \\
% \hline
% HiLight (1e-2) & \textbf{87.63} & \textbf{82.36} & & 86.85 & 69.34 \\
% HiLight (1e-3) &  &  & & 86.89 & \textbf{69.58} \\
% HiLight (1e-6) &  &  & &  &  \\
% \bottomrule
% \end{tabular}
% \caption{Effect of encoding hierarchy without structure encoders. Rec$^\ddagger$ is recursive regularization  \citep{gopal2013recursive} for short.}
% \label{ablate_reg}
% \end{table}

\subsubsection{Label Space Visualization}

In this section, we try to visualize the label space distribution with T-SNE projections. Specifically, we adopt each row of the weight matrix $W$ as the representations of classifiers and project them as well as text representation $h$ into 2D space with T-SNE. As shown in Figure \ref{fig:subfig_1}, the nearest negative labels are the local hard negative labels mined by us. Meanwhile, we can also see that our method can better distinguish the positive labels out of the hard negative labels. Above findings demonstrates the effectiveness of our method.

\section{Conclusion}

In this paper, we propose a Hierarchy-aware Light Global Model with Hierarchical Local ConTrastive Learning (HiLight). We propose a new learning task to introduce the hierarchical information, called Hierarchical Local Contrastive Learning (HiLCL). HiLCL enhances the discriminative ability of classifiers at same path in similar direction by proposed Local Contrastive Learning task (LCL), which is scheduled by Hierarchical Learning (HiLearn). Comparing to previous approaches, our approach achieves competitive results on two benchmark datasets and has the best parameter efficiency and won't collapse at some points. All of the components we designed are proven to be effective.

\section*{Limitations}

Since HiLCL task is text encoder-agnostic, it is easy to apply HiLCL task to a TextRCNN encoder. However, TextRCNN is an old fashion non-pre-trained model and its semantic understanding ability is quite weak. We see an obvious gap between TextRCNN and BERT in previous work \cite{zhu-etal-2023-hitin}. In order to build an efficient model, we do not conduct experiments on TextRCNN.

NYTimes \cite{Sandhaus2008nyt} is also a widely used dataset but due to the business corporation problem, we can't get NYTimes dataset and experiments are not conducted on NYTimes. If we have access to NYTimes dataset in the future, we will verify our model on NYTimes.

% \section*{Acknowledgements}

\bibliography{anthology,custom}

\begin{thebibliography}{22}
\expandafter\ifx\csname natexlab\endcsname\relax\def\natexlab#1{#1}\fi

\bibitem[{Chen(2020)}]{chen2020simple}
et~al. Chen, Ting. 2020.
\newblock A simple framework for contrastive learning of visual
  representations.
\newblock In \emph{International conference on machine learning}, pages
  1597--1607.

\bibitem[{Chen et~al.(2021)Chen, Ma, Lin, and Yan}]{chen-etal-2021-hierarchy}
Haibin Chen, Qianli Ma, Zhenxi Lin, and Jiangyue Yan. 2021.
\newblock \href {https://doi.org/10.18653/v1/2021.acl-long.337}
  {Hierarchy-aware label semantics matching network for hierarchical text
  classification}.
\newblock In \emph{Proceedings of the 59th Annual Meeting of the Association
  for Computational Linguistics and the 11th International Joint Conference on
  Natural Language Processing (Volume 1: Long Papers)}, pages 4370--4379,
  Online. Association for Computational Linguistics.

\bibitem[{Deng et~al.(2021)Deng, Peng, He, Li, and
  Yu}]{deng-etal-2021-htcinfomax}
Zhongfen Deng, Hao Peng, Dongxiao He, Jianxin Li, and Philip Yu. 2021.
\newblock \href {https://doi.org/10.18653/v1/2021.naacl-main.260}
  {{HTCI}nfo{M}ax: A global model for hierarchical text classification via
  information maximization}.
\newblock In \emph{Proceedings of the 2021 Conference of the North American
  Chapter of the Association for Computational Linguistics: Human Language
  Technologies}, pages 3259--3265, Online. Association for Computational
  Linguistics.

\bibitem[{Devlin et~al.(2019)Devlin, Chang, Lee, and
  Toutanova}]{devlin-etal-2019-bert}
Jacob Devlin, Ming-Wei Chang, Kenton Lee, and Kristina Toutanova. 2019.
\newblock \href {https://doi.org/10.18653/v1/N19-1423} {{BERT}: Pre-training of
  deep bidirectional transformers for language understanding}.
\newblock In \emph{Proceedings of the 2019 Conference of the North {A}merican
  Chapter of the Association for Computational Linguistics: Human Language
  Technologies, Volume 1 (Long and Short Papers)}, pages 4171--4186,
  Minneapolis, Minnesota. Association for Computational Linguistics.

\bibitem[{Galanopoulos(2021)}]{Galanopoulos2021Hard}
et~al. Galanopoulos, Damianos. 2021.
\newblock Hard-negatives or non-negatives? a hard-negative selection strategy
  for cross-modal retrieval using the improved marginal ranking loss.
\newblock In \emph{Proceedings of the IEEE/CVF International Conference on
  Computer Vision}, pages 2312--2316.

\bibitem[{Gao(2021)}]{gao2021simcse}
et~al. Gao, Tianyu. 2021.
\newblock Simcse: Simple contrastive learning of sentence embeddings.
\newblock In \emph{Proceedings of the 2021 Conference on Empirical Methods in
  Natural Language Processing}, pages 6894--6910.

\bibitem[{Gopal(2013)}]{gopal2013recursive}
et~al. Gopal, Siddharth. 2013.
\newblock Recursive regularization for large-scale classification with
  hierarchical and graphical dependencies.
\newblock In \emph{Proceedings of the 19th ACM SIGKDD international conference
  on Knowledge discovery and data mining}.

\bibitem[{He(2020)}]{he2020momentum}
et~al. He, Kaiming. 2020.
\newblock Momentum contrast for unsupervised visual representation learning.
\newblock In \emph{Proceedings of the IEEE/CVF conference on computer vision
  and pattern recognition}, pages 9729--9738.

\bibitem[{Jiawei(2019)}]{jiawei2019learning}
et~al. Jiawei, Wu. 2019.
\newblock Learning to learn and predict: A metalearning approach for
  multi-label classification.
\newblock In \emph{Proceedings of the 2019 Conference on Empirical Methods in
  Natural Language Processing and the 9th International Joint Conference on
  Natural Language Processing (EMNLP-IJCNLP)}, page 4354–4364, Hong Kong,
  China. Association for Computational Linguistics.

\bibitem[{Jie et~al.(2022)Jie, Li, and Lu}]{jie-etal-2022-learning}
Zhanming Jie, Jierui Li, and Wei Lu. 2022.
\newblock \href {https://doi.org/10.18653/v1/2022.acl-long.410} {Learning to
  reason deductively: Math word problem solving as complex relation
  extraction}.
\newblock In \emph{Proceedings of the 60th Annual Meeting of the Association
  for Computational Linguistics (Volume 1: Long Papers)}, pages 5944--5955,
  Dublin, Ireland. Association for Computational Linguistics.

\bibitem[{Kim(2021)}]{kim2021self}
et~al. Kim, Taeuk. 2021.
\newblock Self-guided contrastive learning for bert sentence representations.
\newblock In \emph{Proceedings of the 59th Annual Meeting of the Association
  for Computational Linguistics and the 11th International Joint Conference on
  Natural Language Processing (Volume 1: Long Papers)}, pages 2528--2540.

\bibitem[{Kowsari(2017)}]{kowsari2017hdltex}
et~al. Kowsari, Kamran. 2017.
\newblock Hdltex: Hierarchical deep learning for text classification.
\newblock In \emph{2017 16th IEEE international conference on machine learning
  and applications (ICMLA)}, pages 364--371. IEEE.

\bibitem[{Lai et~al.(2015)Lai, Xu, Liu, and Zhao}]{lai2015recurrent}
Siwei Lai, Liheng Xu, Kang Liu, and Jun Zhao. 2015.
\newblock Recurrent convolutional neural networks for text classification.
\newblock In \emph{Proceedings of the AAAI conference on artificial
  intelligence}, volume~29.

\bibitem[{Lewis(2004)}]{lewis2004rcv1}
et~al. Lewis, David~D. 2004.
\newblock Rcv1: A new benchmark collection for text categorization research.
\newblock \emph{Journal of machine learning research}, 5(Apr):361--397.

\bibitem[{Mao(2021)}]{mao2021boosting}
et~al. Mao, Xin. 2021.
\newblock Boosting the speed of entity alignment 10×: Dual attention matching
  network with normalized hard sample mining.
\newblock In \emph{Proceedings of the Web Conference}, pages 821--832.

\bibitem[{Mikolov(2013)}]{mikolov2013distributed}
et~al. Mikolov, Tomas. 2013.
\newblock Distributed representations of words and phrases and their
  compositionality.
\newblock \emph{Advances in neural information processing systems 26}.

\bibitem[{Sandhaus(2008)}]{Sandhaus2008nyt}
Evan. Sandhaus. 2008.
\newblock \emph{The new york times annotated corpus}.
\newblock Philadelphia: Linguistic Data Consortium.

\bibitem[{Thomas N.~Kipf(2017)}]{Thomas2017Semisupervised}
et~al. Thomas N.~Kipf. 2017.
\newblock Semisupervised classification with graph convolutional networks.
\newblock In \emph{5th International Conference on Learning Representations
  Conference Track Proceedings, ICLR 2017}, Toulon, France, April 24-26.

\bibitem[{Wang et~al.(2022)Wang, Wang, Huang, Sun, and
  Wang}]{wang-etal-2022-incorporating}
Zihan Wang, Peiyi Wang, Lianzhe Huang, Xin Sun, and Houfeng Wang. 2022.
\newblock \href {https://doi.org/10.18653/v1/2022.acl-long.491} {Incorporating
  hierarchy into text encoder: a contrastive learning approach for hierarchical
  text classification}.
\newblock In \emph{Proceedings of the 60th Annual Meeting of the Association
  for Computational Linguistics (Volume 1: Long Papers)}, pages 7109--7119,
  Dublin, Ireland. Association for Computational Linguistics.

\bibitem[{Yuning~Mao(2019)}]{yuning2019hierarchical}
et~al. Yuning~Mao. 2019.
\newblock Hierarchical text classification with reinforced label assignment.
\newblock In \emph{Proceedings of the 2019 Conference on Empirical Methods in
  Natural Language Processing and the 9th International Joint Conference on
  Natural Language Processing (EMNLP-IJCNLP)}, page 445–455, Hong Kong,
  China. Association for Computational Linguistics.

\bibitem[{Zhou(2020)}]{jie2020hierarchy}
et~al. Zhou, Jie. 2020.
\newblock Hierarchy-aware global model for hierarchical text classification.
\newblock In \emph{Proceedings of the 58th Annual Meeting of the Association
  for Computational Linguistics (ACL)}.

\bibitem[{Zhu(2023)}]{zhu-etal-2023-hitin}
et~al. Zhu, He. 2023.
\newblock \href {https://aclanthology.org/2023.acl-long.432} {{H}i{TIN}:
  Hierarchy-aware tree isomorphism network for hierarchical text
  classification}.
\newblock In \emph{Proceedings of the 61st Annual Meeting of the Association
  for Computational Linguistics (Volume 1: Long Papers)}, pages 7809--7821,
  Toronto, Canada. Association for Computational Linguistics.

\end{thebibliography}
\bibliographystyle{acl_natbib}

\appendix

% \section{Example Appendix}
% \label{sec:appendix}

\end{document}